\documentclass[lettersize,journal]{IEEEtran}
\usepackage{amsmath,amsfonts}
\usepackage{algorithmic}
\usepackage{algorithm}
\usepackage{array}
\usepackage[caption=false,font=normalsize,labelfont=sf,textfont=sf]{subfig}
\usepackage{textcomp}
\usepackage{stfloats}
\usepackage{url}
\usepackage{verbatim}
\usepackage{graphicx}
\usepackage{cite}
\usepackage{enumitem}
\usepackage{multirow}

\usepackage{arydshln}

\usepackage{caption2}
\usepackage{graphicx}
\usepackage{wrapfig}
\usepackage{bbding}
\usepackage{amsmath}
\usepackage{wrapfig}
\usepackage{booktabs}
\usepackage{bbm}
\usepackage{dsfont}
\usepackage[table]{xcolor}

\providecommand{\onedot}{.}
\usepackage{xspace} 
\newcommand{\eg}{\textit{e.g}\onedot\xspace}

\usepackage[switch]{lineno}
\begin{document}

\title{AV-Unified: A Unified Framework for Audio-visual Scene Understanding}

\author{
Guangyao Li, 
Xin Wang,~\IEEEmembership{IEEE Member}, 
Wenwu Zhu,~\IEEEmembership{IEEE Fellow}

% <-this % stops a space
% \thanks{This paper was produced by the IEEE Publication Technology Group. They are in Piscataway, NJ.}
% <-this % stops a space
% \thanks{Manuscript received April 19, 2024; revised August 16, 2024.}

\thanks{
Guangyao Li, Xin Wang and Wenwu Zhu are with the Department of Computer Science and Technology, Tsinghua University, Beijing 100084, China, and also with the Beijing National Research Center for Information Science and Technology, Beijing 100084, China.
E-mail: \{guangyaoli, xin\_wang, wwzhu\}@tsinghua.edu.cn.
Corresponding authors: Xin Wang and Wenwu Zhu.
The project supported by the National Natural Science Foundation of China (No. 62502268, 62222209), the China Postdoctoral Science Foundation (No.2024M761681), and Beijing National Research Center for Information Science and Technology under Grant No.BNR2023TD03006.
}

}

% \markboth{Journal of \LaTeX\ Class Files,~Vol.~14, No.~8, August~2025}%
\markboth{SUBMITTED TO IEEE TRANSACTIONS ON MULTIMEDIA}%
{Shell \MakeLowercase{\textit{et al.}}: A Sample Article Using IEEEtran.cls for IEEE Journals}

% \IEEEpubid{0000--0000/00\$00.00~\copyright~2025 IEEE}

\maketitle

\begin{abstract}
When humans perceive the world, they naturally integrate multiple audio-visual tasks within dynamic, real-world scenes. However, current works such as event localization, parsing, segmentation and question answering are mostly explored individually, making it challenging to comprehensively understand complex audio-visual scenes and explore inter-task relationships. Hence, we propose \textbf{AV-Unified}, a unified framework that enables joint learning across a wide range of audio-visual scene understanding tasks. AV-Unified standardizes the diverse input-output formats of each task and incorporates a multi-scale spatiotemporal perception network to effectively capture audio-visual associations. Specifically, we unify the inputs and outputs of all supported tasks by converting them into sequences of discrete tokens, establishing a shared representation that allows a single architecture to be trained jointly across heterogeneous varied datasets. Considering the varying temporal granularity of audio-visual events, a multi-scale temporal perception module is designed to capture key cues. Meanwhile, to overcome the lack of auditory supervision in the visual domain, we design a cross-modal guidance-based spatial perception module that models spatial audio-visual associations. Furthermore, task-specific text prompts are employed to enhance the model’s adaptability and task-awareness. Extensive experiments on benchmark datasets (\eg, AVE, LLP, MUSIC-AVQA, VGG-SS and AVS) demonstrate the effectiveness of AV-Unified across temporal, spatial, and spatiotemporal tasks.

\end{abstract}


\begin{figure*}[t]
     \centering
     \includegraphics[width=1\textwidth]{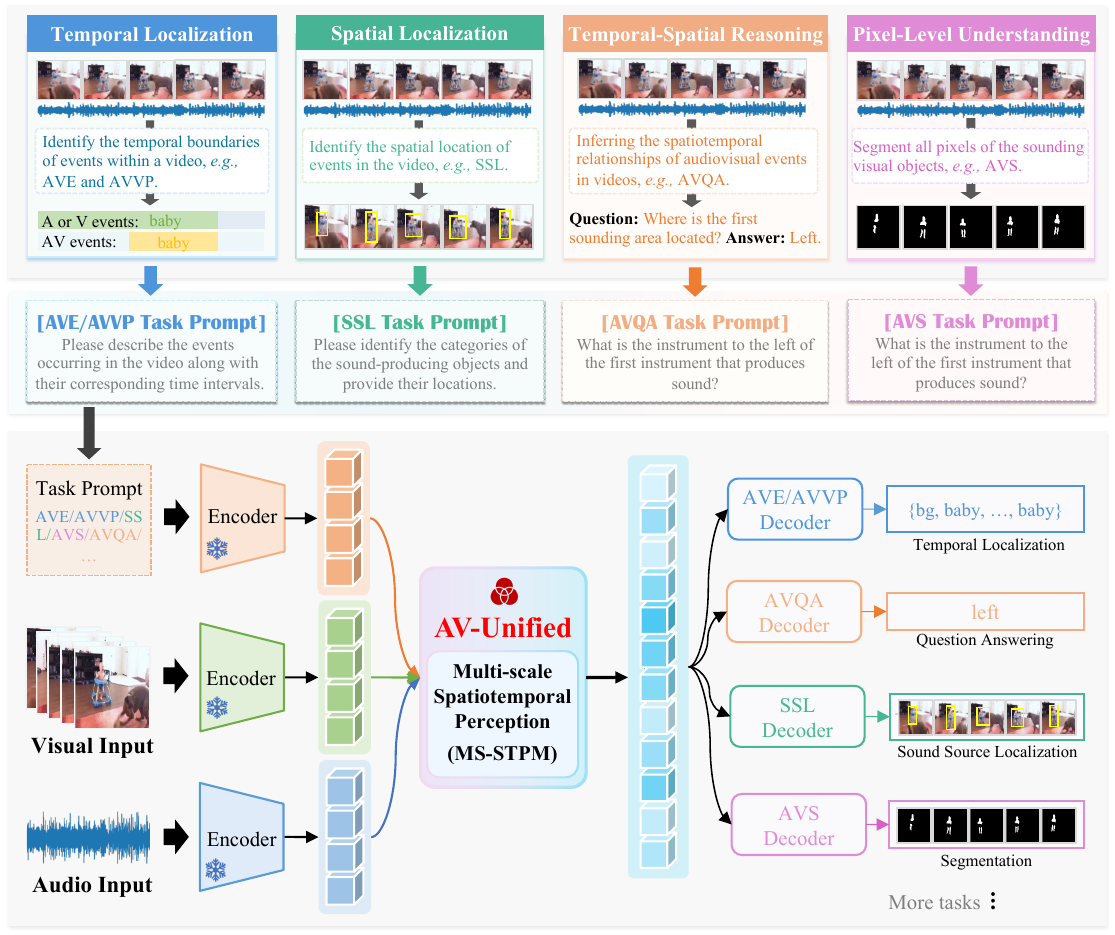}
     \vspace{-1.6em}
     \caption{AV-Unified is a single sequence-to-sequence model that performs a variety of audio-visual tasks using a unified architecture without a need for either task or modality specific branches. A schematic of the model with multiple demonstrative audio-visual tasks: event localization, video parsing, sound source localization, segmentation and question answering.
     }
     \label{fig:teaser}
    \vspace{-1em}
\end{figure*}

\section{Introduction}
Cognitive neuroscience suggests that when multiple sensory modalities such as auditory and visual systems work together, the brain forms a highly consistent and enriched perception of the environment~\cite{holmes2005multisensory, wang2024multi}. This integration enables cross-modal information transfer and understanding, allowing the brain to transcend individual sensory boundaries. 
Such multimodal perception and reasoning are fundamental aspects of cognitive intelligence and are essential for human comprehension of the world. Vision and hearing, as the two most important senses for humans to perceive the world, describe objects of interest from different perspectives and are often complementary. By utilizing the audio-visual elements within these multimodal scenes, it is possible to explore more comprehensive scene information~\cite{wei2022learning}, thereby overcoming the limitations of perception restricted to a single modality.

In recent years, significant progress has been made in leveraging audio and visual modalities jointly to enhance the understanding of multimodal scenes, such as event localization (AVE)~\cite{tian2018audio, brousmiche2021multi}, video parsing (AVVP)~\cite{tian2020unified, rachavarapu2023boosting}, sound source localization (SSL)~\cite{senocak2018learning, hu2021class}, segmentation (AVS)~\cite{zhou2022audio, liu2023annotation, li2023catr}, and question answering (AVQA)~\cite{yun2021pano, li2022learning, yang2022avqa}, \textit{etc}. 
Among these, AVE~\cite{tian2018audio} and AVVP~\cite{tian2020unified} focus on localizing the temporal boundaries of visual and auditory events; SSL~\cite{senocak2018learning} localizes sound sources by learning the co-occurrence of audio and visual cues; AVS~\cite{zhou2022audio} seeks to accurately segment the complete appearance of objects making sound within video frames; and AVQA~\cite{li2022learning} aim to answer questions related to different visual objects, sounds, and their associations in videos. While these studies have made significant progress, they mostly have been explored based on individual tasks, making it challenging to comprehensively understand complex audio-visual scenes. Considering that humans do not dissect scenes into multiple subtasks when perceiving the world, but instead have a unified understanding of multiple tasks within a scene, it's evident that these tasks are often interrelated and can mutually assist each other. 
For instance, when a baby and a dog are playing, a caregiver can simultaneously perceive the baby's laughter and the spatial location of the laughter in the field of view, thus better understanding the baby's emotions. Hence, unifying multiple audio-visual tasks in a single framework, allowing a shared-parameter model to solve them, is a valuable topic.

Recently, successful explorations in task unification in the fields of Computer Vision (CV) and Natural Language Processing (NLP) have shown the potential for unifying audio-visual scene understanding tasks. Existing research has attempted to unify tasks in audio-visual learning, such as ONEAVM~\cite{mo2023unified}, which addresses tasks like sound source separation and localization but is essentially a multi-task learning paradigm. UniAV~\cite{geng2024uniav} has only unified temporal localization tasks, neglecting the importance of spatial localization, highlighting the limitations in the comprehensiveness of existing research. More recently, Crab~\cite{du2025crab} has achieved initial progress in multi-task joint training, but it relies on fine-tuning with externally constructed data. Thus, developing a unified framework that integrates temporal localization tasks (\eg, AVE~\cite{tian2018audio} and AVVP~\cite{tian2020unified}) with spatial localization tasks (\eg, SSL~\cite{senocak2018learning} and AVS~\cite{zhou2022audio}) and extends to spatiotemporal reasoning tasks (\eg, AVQA~\cite{li2022learning}) is crucial for advancing comprehensive audio-visual scene understanding.

To achieve the goal of a unified framework for audio-visual scene understanding tasks, several challenges need to be addressed. \textbf{Firstly}, these tasks encompasses various aspects, including temporal event localization, spatial segmentation, and spatiotemporal reasoning. It involves multimodal data such as images, audio, text, and masks. Thus, a key challenge lies in integrating these diverse modalities into a unified input format while ensuring the model generates consistent outputs adaptable to various downstream tasks. \textbf{Secondly}, unifying audio-visual scene understanding tasks is inherently more complex than handling single-modal visual or language tasks, introducing unique difficulties:
1) Temporal perception: Previous works~\cite{tian2018audio, tian2020unified, zhou2022audio, li2022learning} typically samples events uniformly per second, which can disrupt the continuity of events.  In real-world scenes, events in natural scenes may span multiple seconds and occur at varying time scales. Hence, it is essential to design models capable of capturing multi-scale audio-visual events to accurately reflect their completeness and continuity.
2) Spatial perception: The lack of supervised information for spatial visual objects and sounds 
makes it difficult for models to associate sounds with corresponding visual regions in videos.
While some existing studies leverage pre-trained object detectors to identify salient regions,
these models are typically trained on datasets that lack rare or domain-specific object categories in audio-visual scene (\eg, suona, erhu in AVQA task).
This limitation reduces their effectiveness in locating relevant targets, thereby hindering accurate sound-region association and overall spatial perception.

To track these challenges, we propose a Unified Audio-Visual Perception Framework (\textbf{AV-Unified}) that enables joint learning across multiple audio-visual scene understanding tasks, as illustrated in Fig~\ref{fig:teaser}. \textbf{In terms of task paradigm}, AV-Unified unifies the diverse input and output formats of various tasks into a unified sequence-to-sequence paradigm, allowing all tasks to be trained using a single model with shared parameters. To effectively support both temporal and spatial perception within this multi-task framework, we design a \textbf{M}ulti-scale \textbf{S}patio\textbf{t}emporal \textbf{P}erception \textbf{M}odel (\textbf{MS-STPM}) that extracts unified representations adaptable to all tasks. 
Additionally, we employ a task-prompt guided learning module that enables the model to automatically attend to task-specific representational cues. Specifically, 
\textbf{For temporal perception}, considering the varying durations of audio-visual events,  we designed a multi-scale temporal perception module to integrate audio and visual events at different temporal scales, capturing multimodal information over various time spans.
\textbf{For spatial perception}, to effectively model the spatial association between audio and visual modalities, a cross-modal guidance-based spatial perception module is developed where audio and visual modalities guide each other, facilitating alignment between visual patch sequences and audio signals, and supporting the learning of fine-grained spatial associations in complex scenes. \textbf{For task-specific prompt}, given that different tasks rely on distinct spatiotemporal information, the model needs to selectively attend to relevant features. To this end, we design a prompt-guided feature selection module, which enhances the model’s ability to adaptively extract and emphasize the most informative representations for each task. Extensive experiments on benchmark datasets including AVE, LLP, MUSIC-AVQA, VGG-SS and AVS validate the effectiveness of AV-Unified across temporal, spatial, and spatiotemporal tasks.

Our contributions can be summarized as follows:
\begin{itemize}[leftmargin=*]
\item Unified the inputs and outputs of classic audio-visual scene understanding tasks (\textit{temporal localization: AVE, AVVP; spatial localization: SSL; pixel-level understanding: AVS; spatiotemporal reasoning: AVQA}) by converting all tasks into a sequence-to-sequence format and training them through a shared parameter network.

\item A multi-scale spatiotemporal perception model is proposed to capture events at varying scales and effectively establish audio-visual associations within spatial contexts, addressing the challenges of diverse event durations and the lack of supervisory signals for spatial audio-visual alignment.

\item The proposed task-prompt guided learning module is introduced to steer the model toward learning representations that are specifically relevant to each task.

\item The proposed AV-Unified achieves well performance across multiple tasks on several benchmark datasets, thoroughly demonstrating its effectiveness and versatility in comprehensive audio-visual scene understanding.
\end{itemize}

\section{Related Works}
\subsection{Audio-visual Scene Understanding}
Inspired by the multisensory perception of humans, the community has paid more and more attention to audio-visual scene understanding in recent years~\cite{wei2022learning, chen2024real, mahmud2024ma, duan2023cross, cheng2024mixtures, wang2024ref, gong2025avs, gong2025complementary, yan2024referred, li2023multi, guo2025audio, zhou2025mettle, liavqacot, nie2024dynamic, chen2024verified, huang2024vtimellm, zhou2023intra, wang2023mixup, li2024realtcd}. Visual and auditory modalities have distinct features~\cite{wang2023disentangled,wang2023tiva,chen2023curriculum,qian2022dynamic, niu2022efficient,ding2023image}, including cognitive grounding, semantic and spatiotemporal consistency, and strong support from real-world data. It includes various interesting tasks such as event localization (AVE)~\cite{tian2018audio, brousmiche2021multi}, video parsing (AVVP)~\cite{tian2020unified, mo2022multimodal, hou2023towards}, sound source localization (SSL)~\cite{senocak2018learning, hu2021class}, segmentation (AVS)~\cite{zhou2022audio, liu2023annotation, li2023catr, wang2024prompting}, question answering (AVQA)~\cite{li2022learning, li2024boosting, li2025patch}, \textit{etc}. AVE~\cite{tian2018audio} aims to identify auditory and sound events within a video. Addressing the limitations of audiovisual event localization, Tian~\cite{tian2020unified} further introduced the AVVP task, which aims to localize the temporal boundaries of events in a video and categorize them as audible, visible, or both, for a more granular scene understanding. The SSL~\cite{senocak2018learning, hu2021class} aims to semantically associate sounds with corresponding visual regions without relying on category annotations. It requires both localizing the sounding object in the visual scene and identifying its category. AVS~\cite{zhou2022audio} aims to accurately segment the complete appearance of objects producing sound in video frames, using audio as a guiding signal to determine which object to segment and obtain its complete pixel mask. AVQA~\cite{li2022learning} aims to answer questions related to different visual objects, sounds, and their associations in the video. These studies integrate rich audiovisual cues within multimodal scenes to overcome limitations in perception inherent to single modalities, thereby utilizing both auditory and visual modalities to explore finer-grained scene comprehension.

Apart from the above methods that facilitate scene understanding by excavating and analyzing different modalities, a unified model should be able to perception their spatio-temporal correlation. Hence, the AV-Unified framework is proposed, which achieves joint learning of the AVE, AVVP, SSL, AVS and AVQA tasks.

\begin{figure*}[t]
     \centering
     \includegraphics[width=1\textwidth]{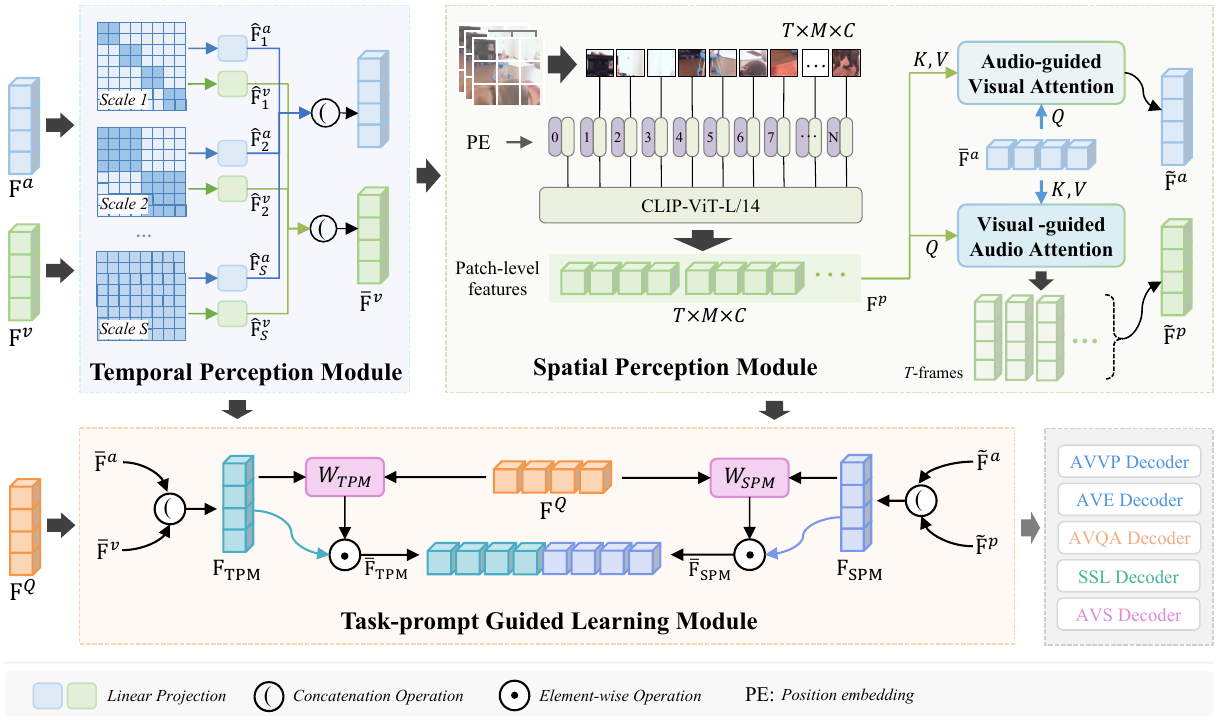}
     \vspace{-1em}
     \caption{The proposed Multi-scale Temporal-Spatial Perception Framework. First, the visual and audio features extracted by the encoder are fed into a temporal perception module to capture key audio-visual temporal cues. Then, a spatial perception module performs cross-modal guidance and interaction based on these temporal cues, uncovering spatial associations between the audio and visual modalities. Next, carefully designed task-specific textual prompts guide the model to focus on features that are most relevant to the current task. Finally, the learned representations are serialized and passed to task-specific decoders to address different downstream audiovisual scene understanding tasks.
     }
     \label{fig:pipe}
    \vspace{-1em}
\end{figure*}

\subsection{Audio-Visual Unified Framework}

Multimedia scene understanding tasks often involve diverse input-output formats, including images, video frames, and pixel-level masks. In recent years, there has been growing interest in developing unified model architectures to enable more generalized and scalable scene understanding. A widely adopted technical approach is to standardize the representation of task inputs and outputs, typically through tokenization~\cite{guo2022toward,tan2022bidirectional,guo2019nat, zhang2025pede}. This enables a consistent modeling interface across different tasks, as illustrated by UniTab~\cite{yang2022unitab}, Pix2Seq~\cite{chen2022unified}, and Unified-IO~\cite{lu2022unified}. Recently, some studies have begun incorporating the audio modality into unified models. For instance, ONE-AVM~\cite{mo2023unified} unifies localization, separation, and recognition within a single methodological framework, though it fundamentally follows a multi-task learning paradigm. Unified-IO2~\cite{lu2024unified} includes the audio modality but lacks explicit modeling of cross-modal audiovisual associations. UniAV~\cite{geng2024uniav} focuses solely on temporal tasks, while Meerkat~\cite{chowdhury2024meerkat} considers both temporal and spatial tasks but simplifies audio-visual segmentation by converting pixel-level masks into bounding boxes, thereby lacking fine-grained pixel-level understanding. The recent Crab~\cite{du2025crab} has made initial progress in jointly training multiple tasks, but it requires fine-tuning with externally constructed data. These efforts highlight the progress made toward unified learning frameworks. However, most existing approaches still underutilize the audio modality and, more critically, fail to model the intrinsic relationships between audio and visual modalities in a unified manner.

In contrast to previous work, the AV-Unified framework proposed in this paper not only unifies the input-output formats of multiple audio-visual tasks but also explicitly and systematically models the deep correlations between audio and visual modalities, enabling a more comprehensive understanding of complex audio-visual scenes.

\section{AV-Unified Framework}

We proposes a unified framework (AV-Unified) that achieves joint learning for multiple audiovisual scene understanding tasks. AV-Unified standardizes the diverse input and output formats of each task and integrates a \textbf{M}ulti-\textbf{s}cale \textbf{S}patio\textbf{t}emporal \textbf{P}erception \textbf{M}odel (\textbf{MS-STPM}) for modeling audio-visual associations, as illustrated in Fig.~\ref{fig:pipe}.

\vspace{-0.5em}
\subsection{Unified Task Representations}
Given an input video sequence containing both visual and audio tracks, we divide it into $T$ non-overlapping audio and visual segments pairs $\{a_t, v_t\}_{t=1}^T$, where each segment spans one second. Subsequently, each video frame is further divided into $M$ patches, and a special $\mathtt{[CLS]}$ token  is prepended to the first patch. For the given task-specific text prompt $Q$, we tokenize it into $N$ individual words, denoted as $\{q_n\}_{n=1}^N$.

\textbf{Audio representation}.
For each audio segment $a_t$, using the pre-trained VGGish~\cite{gemmeke2017audio} model to extract its feature, denoted as $f^a_t \in \mathbb{R}^{D}$, where $D$ is the feature dimension. The VGGish model is a VGG-like 2D CNN pre-trained on the large-scale AudioSet~\cite{gemmeke2017audio} dataset, operating on transformed audio spectrograms. The resulting second-level spectrogram features over time can be represented as $F^a = \{f^a_1, f^a_2, \ldots, f^a_T\}$.

\textbf{Visual representation}.
A fixed number of frames are sampled from each visual segment $v_t$. Then we apply pre-trained CLIP~\cite{radford2021learning}, with frozen parameters, extract both frame-level and token-level features as $f^v_t$ and $f^p_t$ on video frames, respectively, where $f^v_t \in \mathbb{R}^{D}$, $f^p_t \in \mathbb{R}^{M \times D}$ and $M$ are token numbers of one frame.
Finally, the visual frame-level and token-level features can be denoted as $F_v = \{f^v_1, f^v_2, ..., f^v_T\}$, $F_{p} = \{f^p_1, f^p_2, ..., f^p_T\}$, respectively.

\textbf{Text representation}.
Given the task-specific text prompt $Q$, we represent each word $q_n$ in a fixed length vector with word embeddings, and then feed it into the pre-trained CLIP\cite{radford2021learning} model to get the text feature $F^Q$, where $F^Q\in\mathbb{R}^{D}$. And the first $[CLS]$ token is used for extracting text features. Note that for the AVQA task, the text prompt is the asked question.

\vspace{-0.5em}
\subsection{Temporal Perception Module}
To effectively model audio-visual events with varying durations and temporal scales, we propose a Multi-Scale Temporal Perception Module (\textbf{TPM}) designed to capture both fine-grained and coarse-grained temporal dependencies across audio and visual modalities. Specifically, we introduce a multi-scale window attention mechanism with varying window sizes, such as a stacked shifted-window Transformer, where the window size increases progressively with network depth. To emphasize the importance of local context, our attention strategy employs multi-scale window attention centered around each token. Given a fixed window size $S$ each token attends to $S/2$ neighboring tokens on both sides. In each scale, given a set of audio-visual features $\mathbf{F}^a = \{\mathbf{f}^a_t\}^T_{t=1}$, $\mathbf{F}^v =\{\mathbf{f}^v_t\}^T_{t=1}$ in $T$ segments, HAN~\cite{tian2020unified} applied self-attention and cross-attention layers to aggregate the unimodal and cross-modal information at each timestamp:
\begin{align}
\label{agg_a}
\hat{\mathbf{F}}^a_{i} = \phi_{sa}(\mathbf{f}^a_{i, t}, \mathbf{F}^a_i),  
+
\phi_{ca}(\mathbf{f}^a_{i, t}, \mathbf{F}^v_i),
i = 2, 4, 6, ..., S,
\\
\hat{\mathbf{F}}^v_{i} = \phi_{sa}(\mathbf{f}^v_{i, t}, \mathbf{F}^v_i),  
+
\phi_{ca}(\mathbf{f}^v_{i, t}, \mathbf{F}^a_i),
i = 2, 4, 6, ..., S,
\end{align} 

\noindent 
where $S$ is the size of the sliding window. 
Then the transformer encoder is employed to aggregate both within-modality and cross-modality information using multi-head attention blocks:
\begin{footnotesize}
\begin{align}
\label{self_attn}
\phi_{sa}(\mathbf{f}^a_{i, t}, \mathbf{F}^a_i) = \mathcal{G}(\frac{\mathbf{f}^a_{i, t} {\mathbf{F}^a_i}^{\top }}{\sqrt{d}}) \mathbf{F}^a_i,
% \quad
\phi_{ca}(\mathbf{f}^a_{i, t}, \mathbf{F}^v_i) = \mathcal{G}(\frac{\mathbf{f}^a_{i, t} {\mathbf{F}^v_i}^{\top }}{\sqrt{d}}) \mathbf{F}^v_i,
\end{align}
\end{footnotesize}

\noindent 
where $\mathcal{G}(\cdot)$ denotes the \textit{Softmax} function, and $\phi_{sa}(\cdot)$ and $\phi_{ca}(\cdot)$ represent the self-attention and cross-attention operations, respectively. These operations apply dot-product attention over features across temporal steps using non-shared MLPs. Then, we aggregate $\hat{\mathbf{f}}^a_{s, t}$ and $\hat{\mathbf{f}}^v_{s, t}$ across all stages:
\begin{footnotesize}
\begin{align}
\label{self_attn}
\overline{\mathbf{F}}^a = 
\boldsymbol{\Phi}([\hat{\mathbf{F}}^a, \hat{\mathbf{F}}^a_1, 
\hat{\mathbf{F}}^a_2, 
..., \hat{\mathbf{F}}^a_S]),
\quad
\overline{\mathbf{F}}^v = 
\boldsymbol{\Phi}([\hat{\mathbf{F}}^v, \hat{\mathbf{F}}^v_1, 
\hat{\mathbf{F}}^v_2, 
..., \hat{\mathbf{F}}^v_S]),
\end{align}
\end{footnotesize}

\noindent 
where $\hat{\mathbf{F}}^a_S = \{\hat{\mathbf{f}}^a_{s, 1}, \hat{\mathbf{f}}^a_{s, 2}, ..., \hat{\mathbf{f}}^a_{s, t}\}$, $\hat{\mathbf{F}}^v_S = \{\hat{\mathbf{f}}^v_{s, 1}, \hat{\mathbf{f}}^v_{s, 2}, ..., \hat{\mathbf{f}}^v_{s, t}\}$, and $\boldsymbol{\Phi}$ is concatenate operation. With the proposed TPM, the model is able to capture critical audio-visual cues along the temporal dimension for improved contextual understanding.

\subsection{Spatial Perception Module}
Considering that the position of sound sources and their corresponding visual objects often reflects the spatial correlation between audio and visual modalities, we propose a cross-modal guidance-based Spatial Perception Module (\textbf{SPM}). This module leverages strong cross-modal perception capabilities to effectively model spatial associations between audio-visual semantic signals. Specifically, at each time step, there are significant spatial correlations among image patches within a video frame. Given the visual patch-level features $\mathbf{F}^p$ and audio embedding $\mathbf{F}^a$, we first apply a self-attention mechanism to model the intra-frame relationships among visual patches, thereby enhancing the patch-level representation of each video frame. This process can be formally expressed as:
\begin{align}
\label{self_attn}
\hat{\mathbf{f}}^p_{t, m} = \mathbf{f}^p_{t, m} + 
\phi_{sa}(\mathbf{f}^p_{t, m}, \mathbf{F}^p_t),
\end{align}
where $m=\{1, 2, ..., M\}$.
To obtain semantically aligned cross-modal representations, we subsequently perform bi-directional cross-modal attention, where audio features guide the refinement of visual representations and vice versa. Specifically, we apply audio-guided visual attention and visual-guided audio attention to produce enhanced modality-specific features. This process is formally defined as:
\begin{align}
\label{self_attn}
\widetilde{\mathbf{f}}^p_{t, m} = \hat{\mathbf{f}}^p_{t, m} +
\phi_{ca}(\hat{\mathbf{f}}^p_{t, m}, \mathbf{f}^a_t),
\quad
\widetilde{\mathbf{f}}^a_t = \overline{\mathbf{f}}^a_t +
\phi_{ca}(\overline{\mathbf{f}}^a_t, \hat{\mathbf{F}}^p_t),
\end{align} 
where $\hat{\mathbf{f}}^p_{t, m} \in \mathbb{R}^{M \times C}$, and $\hat{\mathbf{F}}^p_t = \{\hat{\mathbf{f}}^p_{t, 1}, \hat{\mathbf{f}}^p_{t, 2},..., \hat{\mathbf{f}}^p_{t, M} \}$, $\hat{\mathbf{F}}^p_t \in \mathbb{R}^{M \times C}$.
Then, we aggragegate $\widetilde{\mathbf{f}}^a_t$, $\widetilde{\mathbf{f}^p_t} = \{\widetilde{\mathbf{f}}^p_{t, m}\}_{m=1}^M$ in all temporal segments as:
$
\widetilde{\mathbf{F}}^a = \{\widetilde{\mathbf{f}}^a_1, \widetilde{\mathbf{f}}^a_2, ..., \widetilde{\mathbf{f}}^a_T \},
\quad
\widetilde{\mathbf{F}}^p = \{\widetilde{\mathbf{f}^p_1},\widetilde{\mathbf{f}^p_2}, ..., \widetilde{\mathbf{f}^p_T} \},
$
where $\widetilde{\mathbf{F}}^a \in \mathbb{R}^{T \times C}$ and $\widetilde{\mathbf{F}}^p \in \mathbb{R}^{T \times M \times C}$.
$\widetilde{\mathbf{F}}^a$ and $\widetilde{\mathbf{F}}^p$ are the representations obtained from the SMP. With the proposed TPM and SPM, the model is able to establish robust and effective associations between audio-visual cues, thereby enabling it to accurately capture and localize potential sounding regions within video frames.

\begin{figure}[t]
     \centering
     \includegraphics[width=0.475\textwidth]{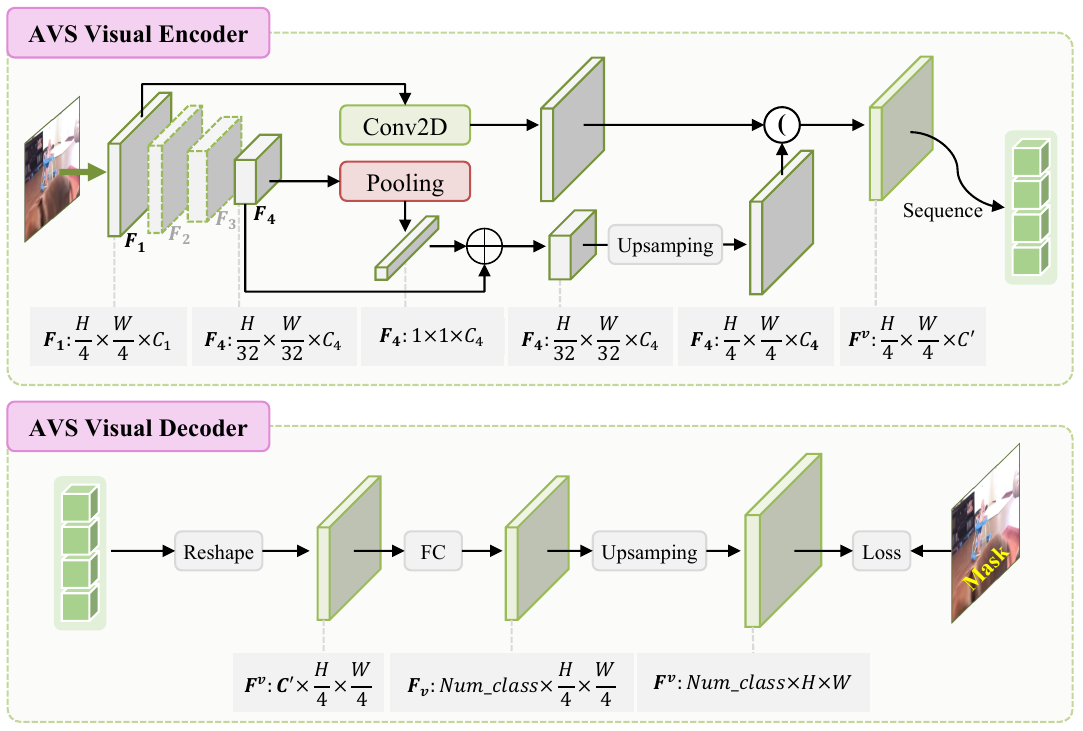}
     \vspace{-0.75em}
     \caption{Encoder and decoder for AVS task.}
     \label{fig:avs}
     \vspace{-1.25em}
\end{figure}

\subsection{Task-prompt Guided Learning Module}
Considering the preferences in training for multiple tasks, such as AVE and AVVP tasks leaning more towards temporal perception, AVS tasks favoring spatial perception, and AVQA task leaning towards spatiotemporal perception. Hence, a Task-prompt Guided Learning Module (\textbf{TPGL}) is designed. Its purpose is to manually set a textual prompt for each task to guide the preferences during training, with the prompt for the AVQA task is input question.

For the given audiovisual representations $\overline{\mathbf{F}}^a, \overline{\mathbf{F}}^p$, and $\widetilde{\mathbf{F}}^a, \widetilde{\mathbf{F}}^p$ obtained through TPM and SPM, we employ a linear projection layer to map their dimensions to the same size and apply a $ReLU$ activation function. Then, for the $\widetilde{\mathbf{F}}^a$ obtained by SPM, we transform its dimension from $T \times C$ to $T \times 1 \times C$ and concatenate it with $\widetilde{\mathbf{F}}^p$ at the patch level. Simultaneously, for $\overline{\mathbf{F}}^a, \overline{\mathbf{F}}^p$ obtained through TPM, we also concatenate them along the temporal dimension using a concatenation operation. This process can be represented as:
\begin{align}
\label{self_attn}
\mathbf{F}_{\mathtt{TPM}} = \boldsymbol{\Phi}([\overline{\mathbf{F}}^a, \overline{\mathbf{F}}^p]),
\quad
\mathbf{F}_{\mathtt{SPM}} = \boldsymbol{\Phi}([\widetilde{\mathbf{F}}^a, \widetilde{\mathbf{F}}^p]),
\end{align}
where 
$\mathbf{F}_{\mathtt{TPM}} \in \mathbb{R}^{2 \times T \times C}$, 
$\mathbf{F}_{\mathtt{SPM}} \in \mathbb{R}^{T \times (M+1) \times C}$.
Then, for the given task-prompt feature $\mathbf{F}^Q \in \mathbb{R}^{1 \times C}$, we use it as the \textit{Query} to calculate similarities separately along the temporal dimension of $\mathbf{F}_{\mathtt{TPM}}$ and the patch dimension of $\mathbf{F}_{\mathtt{SPM}}$:
\begin{footnotesize}
\begin{align}
\label{self_attn}
\mathbf{W}_{\mathtt{TPM}} = \mathcal{G}(\frac{\mathbf{F}^Q {\mathbf{F}_{\mathtt{TPM}}}^{\top}}{\sqrt{d}}) \mathbf{F}_{\mathtt{TPM}},
\mathbf{W}_{\mathtt{SPM}} = \mathcal{G}(\frac{\mathbf{F}^Q {\mathbf{F}_{\mathtt{SPM}}}^{\top}}{\sqrt{d}}) \mathbf{F}_{\mathtt{SPM}},
\end{align}
\end{footnotesize}
where $\mathbf{W}_{\mathtt{TPM}} \in \mathbb{R}^{2T}$, $\mathbf{W}_{\mathtt{SPM}} \in \mathbb{R}^{T \times (M+1)}$. The dimensions of $\mathbf{W}_{\mathtt{TPM}}$ and $\mathbf{W}_{\mathtt{SPM}}$ are transformed into $2T \times 1$ and $T \times (M+1)$ respectively, resulting in updated features $\overline{\mathbf{F}}_{\mathtt{TPM}}$, $\overline{\mathbf{F}}_{\mathtt{SPM}}$:
\begin{align}
\label{self_attn}
\overline{\mathbf{F}}_{\mathtt{TPM}} = 
\mathbf{W}_{\mathtt{TPM}} \odot \mathbf{F}_{\mathtt{TPM}}, 
\quad
\overline{\mathbf{F}}_{\mathtt{SPM}} = 
\mathbf{W}_{\mathtt{SPM}} \odot \mathbf{F}_{\mathtt{SPM}},
\end{align}
where $\overline{\mathbf{F}}_{\mathtt{TPM}} \in \mathbb{R}^{2T \times C}$, $\overline{\mathbf{F}}_{\mathtt{SPM}} \in \mathbb{R}^{T \times C \times (M+1)}$.
Then, they are concatenated into a single sequence to be utilized for various downstream audiovisual tasks.

With the integration of the proposed TPM, SPM, and TPGL, the model is able to effectively capture audio-visual associations in dynamic scenes, enabling a comprehensive understanding of audio-visual environments.

\begin{table*}[t]
\begin{center}
\setlength{\tabcolsep}{9pt}

\caption{The performance of the AV-Unified Framework on the LLP Dataset for the AVVP Task.}
\vspace{0.25em}
\label{tab_avvp}

\scalebox{1}{

\begin{tabular}{c|ccccc|ccccc}
\hline
& \multicolumn{5}{c|}{\textbf{Segment-level}}
& \multicolumn{5}{c}{\textbf{Event-level}}    \\ 
\cline{2-11} 
\multirow{-2}{*}{{ \textbf{Method}}} 
& \textbf{Audio} & \textbf{Visual} & \textbf{Audio-Visual} 
& \textbf{Type} & \textbf{Event} & \textbf{Audio} & \textbf{Visual} & \textbf{Audio-Visual} 
& \textbf{Type} & \textbf{Event} \\ \hline
AVEL~\cite{tian2018audio}    & 47.2     & 37.1      & 35.4         & 39.9    & 41.6     & 40.4     & 34.7      & 31.6        & 35.5    & 36.5     \\
AVSDN~\cite{lin2019dual}        & 47.8     & 52.0      & 37.1         & 45.7    & 50.8     & 34.1     & 46.3      & 26.5        & 35.6    & 37.7     \\
HAN~\cite{tian2020unified}    & 60.1     & 52.9      & 48.9         & 54.0    & 55.4     & 51.3     & 48.9      & 43.0        & 47.7    & 48.0     \\
MA~\cite{wu2021exploring}     & 60.3     & 60.0      & 55.1         & 58.9    & 57.9     & 53.6     & 56.4      & 49.0        & 53.0    & 50.6     \\
CVCMS~\cite{lin2021exploring}        & 60.8     & 63.5      & 57.0         & 60.5    & 59.5     & 53.8     & 58.9      & 49.5        & 54.0    & 52.1     \\
DHHN~\cite{jiang2022dhhn}         & 61.4     & 63.4      & 56.8         & 60.5    & 59.5     & 54.6     & 60.8      & 51.1        & 55.5    & 53.3     \\
MM-Pyramid~\cite{yu2022mm}         & 61.1     & 60.3      & 55.8         & 59.7    & 59.1     & 53.8     & 56.7      & 49.4        & 54.1    & 51.2     \\
MGN~\cite{mo2022multi}    & 60.8     & 55.4      & 50.4         & 55.5    & 57.2     & 51.1     & 52.4      & 44.4        & 49.3    & 49.1     \\

\hline

\rowcolor{gray!10}
\textbf{AV-Unified (Ours)}  &  \textbf{61.6} &\textbf{66.9}    &  \textbf{59.8}      &  \textbf{62.8}       &  \textbf{61.5 } & \textbf{54.8}   & \textbf{63.0}    & \textbf{52.2} &  \textbf{56.7}       &  \textbf{53.9 } \\

\hline
\end{tabular}

}

\vspace{-1.75em}
\end{center}
\end{table*}

\begin{table}{}
\begin{center}
\setlength{\tabcolsep}{9pt} 
\renewcommand\arraystretch{1.1}

\caption{The performance of the AV-Unified on AVE.}
\vspace{0.25em}

\label{tab_ave}

\scalebox{1}{
\begin{tabular}{c|cc}
\hline

\textbf{Method} &   \textbf{Fully-Supervised}  & \textbf{Weakly-Supervised}
\\
\hline
AVEL~\cite{tian2018audio}         & 72.7      & 66.7      \\
AVIN~\cite{ramaswamy2020makes}         & 75.2      & 69.4      \\
AVT~\cite{lin2020audiovisual}          & 75.8      & 70.2      \\
CMRAN~\cite{xu2020cross}        & 77.4      & 73.0      \\
PSP~\cite{zhou2021positive}          & 77.8      & 73.5      \\
MM-Pyramid~\cite{yu2022mm}   & 77.8      & 73.2      \\
\hline

\rowcolor{gray!10}

\textbf{AV-Unified (Ours)} &  \textbf{78.7}      &  \textbf{74.2}  \\ 
\hline
\end{tabular}

}

\vspace{-2.2em}

\end{center}
\end{table}

\subsection{Training Objective}
A strong multimodal model has to be exposed to solving diverse sets of problems during pre-training. To train the AV-Unified multi-task unified framework, we combined several benchmark datasets, including AVE, LLP, VGG-SS, AVS and MUSIC-AVQA. Each training batch is composed of samples from a single task. To mitigate the problem of catastrophic forgetting, we randomly sample a batch from one task in each iteration and update the model parameters using the loss computed for that specific task. Specifically, for a batch of data, assuming that the batch comes from the task \textit{target}, then: 
\begin{align}
\label{self_attn}
LOSS = \sum_{i=1}^{N} w_i \cdot loss_i,
\quad 
\end{align}
where \textit{N} is the total number of tasks, \textit{$loss_i$} represents the loss of the i-th task, $w_i=\mathbbm{1}_{[i=target]}$. It is worth noting that, since the AVS task involves pixel-level masks, both its encoder and decoder adopt the architecture illustrated in Fig.~\ref{fig:avs}.

\section{Experiment and Analysis}

\subsection{Datasets and Evaluation Metrics}

\textbf{Audio-Visual Event (AVE)}~\cite{tian2018audio}.
It contains 4,143 videos covering 28 event categories, \eg, human and animal activities, and vehicle sounds. Videos in AVE dataset are temporally labeled with audio-visual event boundaries. Each video lasts 10 seconds, and each event lasts at least 2 seconds.

\textbf{Look, Listen and Parse (LLP)}~\cite{tian2020unified}.
The LLP consists of 11,849 10-seconds video clips annotated with 25 event categories. It covers various real-life scenes such as speech, music performances, car, cheering, dog, etc. 
We use the 10,000 video clips with only video-level event annotations for model training. The detailed annotations are available for the remaining 1,849 validation and test videos.

\begin{table}{}
\begin{center}
\setlength{\tabcolsep}{20pt} 
\renewcommand\arraystretch{1.1} 
\vspace{-0.3em}
\caption{The performance of the AV-Unified on VGG-SS.}
\vspace{0.25em}
\label{tab_vggss}

\scalebox{1}{
\begin{tabular}{c|c|c}
\hline
\textbf{Method}       & \textbf{CIoU}(\%)                   & \textbf{AUC}(\%) \\ \hline
Attention10K~\cite{senocak2018learning} & \multicolumn{1}{c|}{18.50} & 30.20   \\
CoarsetoFine~\cite{qian2020multiple} & \multicolumn{1}{c|}{29.10} & 34.80   \\
AVObject~\cite{afouras2020self}     & \multicolumn{1}{c|}{29.70} & 35.70   \\
LVS~\cite{chen2021localizing}          & \multicolumn{1}{c|}{34.40} & 38.20   \\
HardPos~\cite{senocak2022learning}      & \multicolumn{1}{c|}{34.60} & 38.00   \\
EZ-VSL~\cite{mo2022localizing}       & \multicolumn{1}{c|}{38.85} & 39.54   \\ 
\hline

\rowcolor{gray!10}  
\textbf{AV-Unified (Ours)}  &  \textbf{39.16}     &  \textbf{41.24}   \\ 
\hline
\end{tabular}

}

\vspace{-2.2em}

\end{center}
\end{table}

\textbf{Audio-visual Segmentation (AVS)}, including \textbf{Semisupervised Single-sound Source Segmentation (S4)}~\cite{zhou2022audio}. It contains a total of 4, 932 videos, with 3,452 videos for training, 740 for validation, and 740 for testing. The target objects cover 23 different categories, including humans, animals, vehicles, and musical instruments. Besides, Each video contains five frames, but only the first frame is annotated. \textbf{Fully-supervised Multiple-sound Source Segmentation (MS3)}~\cite{zhou2022audio}. The MS3 contains 424 videos and each video has multiple sounding sources and the sounding objects are visible in the frames. Each video was trimmed to 5 seconds, covering the same categories as the S4 subset. \textbf{Fully-supervised Audio-Visual Semantic Segmentation (AVSS)}~\cite{zhou2024audio}. The AVSS containing a semantic-labels subset that provides pixel-wise semantic labels, as a significant complement of S4 and MS3. And it includes 12,356 videos covering 70 categories. 

\textbf{VGG Sound Source (VGG-SS)}~\cite{chen2021localizing} 
is designed for evaluating sound source localization. The dataset includes over 200 categories and 5,000 videos, featuring annotated labels based on the VGGSound dataset. Each visible sound source in a video clip is explicitly annotated with bounding boxes.

\textbf{MUSIC-AVQA}~\cite{li2022learning}, 
it contains 9,288 videos covering 22 different musical instruments, with a total duration of over 150 hours and 45,867 QA pairs. The questions are designed under multi-modal scenes containing 33 question templates covering nine types, depending on which modalities are used to discover question-related clues for answer prediction.

\subsection{Implementation Details}

For the visual stream, videos are divided into 1-second segments, with frames sampled at $1fps$. We employ the CLIP-ViT-L-14~\cite{radford2021learning} model pre-trained on ImageNet to extract 512-dimensional feature representations for each visual segment, where the $[\text{CLS}]$ token is used as the visual frame-level feature. For the audio stream, signals are sampled at 16~kHz, a standard sampling rate for audio processing. We use the VGGish network pre-trained on AudioSet to extract 128-dimensional audio features. For each input task-prompt, we adopt the same visual frame-level encoder to extract a 512-dimensional feature vector. All experiments are conducted using the Adam optimizer with an initial learning rate of $1 \times 10^{-4}$, which is decayed by a factor of 0.1 every 10 epochs. The batch size and the total number of training epochs are set to 64 and 200, respectively. The proposed AV-Unified framework is trained on 6$\times$NVIDIA A100-40G GPUs.

\begin{table*}[t]
\begin{center}
\setlength{\tabcolsep}{7.35pt}
\renewcommand\arraystretch{1.1}

\caption{The performance of the AVQA task trained jointly with other multiple audio-visual scene understanding tasks.}

\label{tab_avqa}

\scalebox{0.98}{

\begin{tabular}{c|ccc|ccc|cccccc|c}
\hline
\multirow{2}{*}{\textbf{Method}}    & \multicolumn{3}{c|}{\textbf{Audio}}         & \multicolumn{3}{c|}{\textbf{Visual}}        & \multicolumn{6}{c|}{\textbf{Audio-Visual}}     & \multirow{2}{*}{\textbf{Avg}} \\
      & \textbf{Count}       & \textbf{Comp}        & \textbf{Avg}    & \textbf{Count}       & \textbf{Local}       & \textbf{Avg}    & \textbf{Exist}       & \textbf{Count}       & \textbf{Local}       & \textbf{Comp}        & \textbf{Temp}        & \textbf{Avg}    &       \\ \hline
FCNLSTM~\cite{fayek2020temporal}   & 70.80  & 65.66  & 68.90  & 64.58  & 48.08  & 56.23  & 82.29  & 59.92  & 46.20  & 62.94  & 47.45  & 60.42  & 60.81  \\
GRU~\cite{antol2015vqa}   & 71.29  & 63.13  & 68.28  & 66.08  & 68.08  & 67.09  & 80.67  & 61.03  & 51.74  & 62.85  & 57.79  & 63.03  & 65.03  \\
Hco\_Att~\cite{lu2016hierarchical}  & 70.80  & 54.71  & 64.87  & 63.49  & 67.10  & 65.32  & 79.48  & 59.84  & 48.80  & 56.31  & 56.33  & 60.32  & 62.45  \\
MCAN~\cite{yu2019deep}  & 78.07  & 57.74  & 70.58  & 71.76  & 71.76  & 71.76  & 80.77  & 65.22  & 54.57  & 56.77  & 46.84  & 61.52  & 65.83  \\
PSAC~\cite{li2019beyond}  & 75.02  & 66.84  & 72.00  & 68.00  & 70.78  & 69.41  & 79.76  & 61.66  & 55.22  & 61.13  & 59.85  & 63.60  & 66.62  \\
HME~\cite{fan2019heterogeneous}   & 73.65  & 63.74  & 69.89  & 67.42  & 70.20  & 68.83  & 80.87  & 63.64  & 54.89  & 63.03  & 60.58  & 64.78  & 66.75  \\
HCRN~\cite{le2020hierarchical}  & 71.29  & 50.67  & 63.69  & 65.33  & 64.98  & 65.15  & 54.15  & 53.28  & 41.74  & 51.04  & 46.72  & 49.82  & 56.34  \\
AVSD~\cite{schwartz2019simple}  & 72.47  & 62.46  & 68.78  & 66.00  & 74.53  & 70.31  & 80.77  & 64.03  & 57.93  & 62.85  & 61.07  & 65.44  & 67.32  \\
PanoAVQA~\cite{yun2021pano}  & 75.71  & 65.99  & 72.13  & 70.51  & 75.76  & 73.16  & 82.09  & 65.38  & 61.30  & 63.67  & 62.04  & 66.97  & 69.53  \\
ST-AVQA~\cite{li2022learning}   & 77.78  & 67.17  & 73.87  & 73.52  & 75.27  & 74.40  & 82.49  & 69.88  & 64.24  & 64.67  & 65.82  & 69.53  & 71.59  \\
COCA~\cite{Lao_Pu_Liu_He_Bakker_Lew_2023}  & 79.35  & 67.68 & 75.42  & 75.10  & 75.43  & 75.23  & \textbf{83.50} & 66.63  & 69.72  & 64.12  & 65.57  & 69.96  & 72.33  \\
PSTP-Net~\cite{pstpnet2023li}  & 73.97  & 65.59  & 70.91  & 77.15  & 77.36  & 77.26  & 76.18  & 72.23  & 71.80  & 71.79 & 69.00  & 72.57  & 73.52  \\

TASS~\cite{jiang2025clip} & 
\textbf{83.38} & 63.13 & \textbf{75.92} & \textbf{80.37} & \textbf{79.51} & \textbf{79.93} & 82.39 & 68.91 & 75.89 & 64.40 & 69.22 & 72.33 & 74.98 \\

\hline

\rowcolor{gray!10}
\textbf{AV-Unified (Ours)} & 72.23	& \textbf{78.26}	& 72.60	& 75.93	& 73.38	& 75.61	& 79.00	& \textbf{77.07}	& \textbf{76.27}	& \textbf{76.89}	& \textbf{75.65}	& \textbf{76.96}	& \textbf{76.42}

\\ 
\hline
\end{tabular}
}

\vspace{-1.75em}

\end{center}
\end{table*}

\begin{table}[t]
\begin{center}
\setlength{\tabcolsep}{3.7pt} 

\caption{The performance of the S4, MS3 and AVSS task trained individually and jointly with other multiple audio-visual scene understanding tasks in AV-Unified framework.}

\label{tab_avs}

\scalebox{0.975}{

\begin{tabular}{c|cc|cc|cc}
\hline
\multirow{2}{*}{\textbf{Method}}      & \multicolumn{2}{c|}{\textbf{S4}} & \multicolumn{2}{c|}{\textbf{MS3}} & \multicolumn{2}{c}{\textbf{AVSS}} 
\\
& \textbf{mIoU} & \textbf{F-Score} & \textbf{mIoU}  & \textbf{F-Score} & \textbf{mIoU}  & \textbf{F-Score} \\ \hline
MSSL~\cite{qian2020multiple}        & 44.9    & 66.3     & 26.1     & 36.3     & -    & -    \\
SST~\cite{duke2021sstvos}         & 66.3    & 80.1     & 42.6     & 57.2     & -    & -    \\
iGAN~\cite{mao2021transformer}        & 61.6    & 77.8     & 42.9     & 54.4     & -    &    \\
LGVT~\cite{zhang2021learning}      & 74.9    & 87.3     & 40.7     & 59.3     & -    & -    \\
\hline
TPAVI~\cite{zhou2022audio}         & 78.7    & 87.9     & 54.0     & 64.5     & 29.8     & 35.2     \\
CATR~\cite{li2023catr}        & 81.4    & 89.6     & 59.0     & 67.0     & -    & -    \\
BAVS~\cite{liu2023bavs}        & 82.0    & 88.6     & 58.6     & 65.5     & 32.6     & 36.4     \\
AVSeg~\cite{gao2024avsegformer}         & 82.1    & {\textbf{89.9}}    & 58.4     & \textbf{69.3}    & 36.7     & \textbf{42.0}     \\ 

\hline

\rowcolor{gray!10}
 
\textbf{AV-Unified (Ours)}  &  \textbf{83.2}   &  87.1  &  \textbf{59.5}  & \textbf{69.3}   &  \textbf{37.4}  &  41.9  \\

\hline
\end{tabular}

}

\vspace{-1.5em}
\end{center}
\end{table}

\subsection{Quantitative Results and Analysis}

To validate the effectiveness of the proposed AV-Unified framework, we compare it against recent existing methods across multiple tasks. For temporal localization tasks, we conduct evaluations on the AVE and LLP datasets. For spatial and pixel-level localization tasks, we use the VGG-SS and AVS datasets, respectively. For spatiotemporal reasoning tasks, we perform evaluation on the MUSIC-AVQA dataset.

\textbf{Comparison with other related models.}

As shown in Tab.~\ref{tab_avvp}, \ref{tab_ave}, \ref{tab_vggss}, \ref{tab_avqa}, \ref{tab_avs}, and \ref{tab_single_multi}, the proposed AV-Unified framework consistently improves performance across all tasks. Moreover, the varying difficulty levels of these subtasks place different demands on the model’s spatio-temporal perception capabilities. In joint training, more challenging subtasks tend to enhance the model’s capacity to capture complex spatio-temporal patterns, whereas easier subtasks may dilute this capacity. As a result, the model may perform well on complex tasks but struggle with more simpler ones. For instance, as shown in Tab.~\ref{tab_avqa} and Tab.~\ref{tab_single_multi}, joint training significantly benefits AVQA, particularly in complex reasoning types such as \textit{Counting}, \textit{Localization}, \textit{Comparative}, and \textit{Temporal}, indicating that joint learning helps improve reasoning accuracy on these more complex tasks. However, we also observe slight performance drops under unimodal settings (audio-only or visual-only), suggesting that the current joint optimization strategy leaves room for improvement. Taking the AVS task as an example: AVSS is the most challenging subtask, MS3 represents moderate difficulty, and S4 is the easiest. A comparison of single-task and joint training results in Tab.~\ref{tab_avs} and Tab.~\ref{tab_single_multi} reveals that joint training leads to a drop in F-score for S4, while MS3 shows a 0.9\% improvement. This phenomenon is observed not only across different levels of task complexity but also within the same category of audio-visual scene understanding tasks.

\textbf{Limitation Analysis.}
Comparing the results of unified training with existing methods, we observe a notable performance gap in certain tasks, particularly the S4, AVSS, and MUSIC-AVQA datasets. This discrepancy can be attributed to two main factors: This discrepancy can be attributed to two primary factors. First, many existing methods are task-specific, designed exclusively for a single task without considering cross-task interactions. Such specialization allows for targeted optimization of both model architecture and learning objectives, thereby favoring performance on individual tasks. In contrast, our multi-task unified framework must balance multiple objectives simultaneously, making such task-specific optimization difficult. While multi-task joint learning facilitates the acquisition of shared audiovisual representations, its main advantage lies in adaptability across diverse tasks rather than maximizing performance for any single one.
Second, due to computational resource constraints, we processed the video data for these tasks at a lower sampling rate, which introduced a significant discrepancy in data preprocessing. This difference substantially affects the model’s ability to learn temporal representations, leading to a performance drop compared with existing single-task methods.

\begin{table*}[t]
\begin{center}
\setlength{\tabcolsep}{3.5pt}
\renewcommand\arraystretch{1.15} 

\caption{The performance of multi-task trained individually (AV-Unified \textit{w/o. jt}, \textit{jt}: joint training) and jointly (AV-Unified) with other multiple audio-visual scene understanding tasks in AV-Unified framework.}

\label{tab_single_multi}

\scalebox{0.975}{

\begin{tabular}{c|cc|cccc|cc|cccccc|cc}
\hline
\multirow{3}{*}{\textbf{Method}}   & 
\multicolumn{2}{c|}{AVE~\cite{tian2018audio}}   & 
\multicolumn{4}{c|}{LLP~\cite{tian2020unified}}  & 
\multicolumn{2}{c|}{VGG-SS~\cite{chen2021localizing}}  & 
\multicolumn{2}{c}{S4~\cite{zhou2022audio}}   & 
\multicolumn{2}{c}{MS3~\cite{zhou2022audio}}  & 
\multicolumn{2}{c|}{AVSS~\cite{zhou2024audio}}  & 
\multicolumn{2}{c}{MUSIC-AVQA~\cite{li2022learning}}   \\ 
\cline{2-17} 
& \multirow{2}{*}{Fully} 
& \multirow{2}{*}{Weakly} 
& \multicolumn{2}{c}{Segment-level} 
& \multicolumn{2}{c|}{Event-level} 
& \multirow{2}{*}{CIoU} 
& \multirow{2}{*}{AUC} 
& \multirow{2}{*}{mIoU} 
& \multirow{2}{*}{F-Score} 
& \multirow{2}{*}{mIoU} 
& \multirow{2}{*}{F-Score} 
& \multirow{2}{*}{mIoU} 
& \multirow{2}{*}{F-Score} 
& \multirow{2}{*}{AV} 
& \multirow{2}{*}{All} \\
&  &   & Type  & Event   & Type   & Event   &   &  &   &  &   &  &   &   &   &  \\ 
\hline
AV-Unified \textit{w/o. jt} & 
77.4   & 73.5  & 62.2  & 60.9  & 56.4   & 53.6  & 38.94   & 40.80  & 82.40   & \textbf{89.00}  & 59.30   & 68.40  & 37.10   & 37.40   & 71.94   & 75.32  
\\
\hdashline
\rowcolor{gray!10}
\textbf{AV-Unified (Ours)}
& 
\textbf{78.7} & \textbf{74.2} & \textbf{62.8} & \textbf{61.5}  & \textbf{56.7}   & \textbf{53.9}  & \textbf{39.16}   & \textbf{41.24}  & \textbf{83.20}   & 87.10  & \textbf{59.50}  & \textbf{69.30} & \textbf{43.10}   & \textbf{41.90}  & \textbf{ 76.96 }  & \textbf{ 76.42 } \\ 

\hline
\end{tabular}

}

\vspace{-1.75em}

\end{center}
\end{table*}

\begin{table}{}
\begin{center}
\setlength{\tabcolsep}{9.25pt} 
\renewcommand\arraystretch{1.15} 

\caption{MS-TSPM’s module configuration results.}
\vspace{0.25em}
\label{tab_aba}

\scalebox{0.95}{
\begin{tabular}{c|ccc|c}
\hline
\textbf{Method}  & \textbf{Audio} & \textbf{Visual} & \textbf{Audio-Visual} & \textbf{All}  \\ 
\hline
\textit{w/o} MS-STPM  & 73.93  & 79.23  & 70.37  & 73.35  \\
\textit{w/o} TPM      & \textbf{77.72}  & 79.81  & 71.21  & 74.64  \\
\textit{w/o} SPM      & 75.85  & \textbf{80.64}  & 71.84  & 74.88  \\
\textit{w/o} TPGL     & 75.54  & 79.93  & 71.74  & 74.59  \\ 
\hline

\rowcolor{gray!10}

\textbf{AV-Unified (Ours)} & 72.60 & 75.61  & \textbf{76.96}  & \textbf{76.42} \\ 
\hline
\end{tabular}

}

\vspace{-1.85em}

\end{center}
\end{table}

\vspace{-0.25em}
\subsection{Ablation Studies}

\textbf{Effectiveness of AV-Unified.}
By comparing single-task and unified training results, we observe that training with the unified AV-Unified framework generally yields better performance across most tasks. As shown in Tab.~\ref{tab_avvp} and Tab.~\ref{tab_ave}, for temporal tasks, unified training improves performance on the AVE task by 1.3\% and 0.7\% across two evaluation metrics compared to single-task training. Similarly, for the AVVP task, multiple sub-metrics show performance gains. We also observe improvements in spatial perception tasks. In the multi-source sound segmentation (MS3) subtask shown in Tab.~\ref{tab_avs} and Tab.~\ref{tab_single_multi}, joint training increases mIoU by 0.2\% and F-score by 0.9\%, achieving the best results for this subtask. 
For more challenging spatio-temporal reasoning tasks, AV-Unified improves the average performance metric by 1.10\%. These results demonstrate that incorporating more tasks and training data in a unified framework enhances the model's ability to learn consistent audiovisual representations, thereby benefiting the learning of individual subtasks and validating the effectiveness of the proposed AV-Unified framework.

\textbf{Effectiveness of MS-TSPM.}
To evaluate the impact of MS-TSPM on the AV-Unified framework, we conducted a series of ablation experiments. As shown in Tab.~\ref{tab_aba}, on the MUSIC-AVQA dataset, the model performance drops significantly when the MS-TSPM structure is removed (73.35\% \textit{vs.} 76.42\%).
In addition, by comparing the results of AV-Unified without unified training in Tab.~\ref{tab_single_multi} with those presented in Tab.~\ref{tab_avvp}, \ref{tab_avqa}, \ref{tab_ave}, \ref{tab_vggss}, and \ref{tab_avs}, we observe that even when using MS-TSPM alone, the model still achieves strong performance on specific tasks. This indicates that the model can adapt to different task types and effectively capture the spatiotemporal correlations in audiovisual data, making it well-suited for tasks such as temporal localization, spatial segmentation, and spatiotemporal reasoning. We further analyzed the contribution of each component within MS-TSPM, using the MUSIC-AVQA as a case study. As shown in Tab.~\ref{tab_aba}, when the carefully designed TPM, SPM, and TPGL modules are introduced together, the model achieves the best performance regardless of whether multi-task joint training is applied. Each module also brings a clear performance gain when added individually. In contrast, removing all modules leads to a significant performance drop. These results demonstrate the effectiveness of the proposed components. Working in combination, they help the model better capture the spatiotemporal associations in audiovisual scenes and enhance overall performance.

\vspace{-0.5em}
\subsection{Visualization Results}

\begin{figure}[t]
     \centering
     \includegraphics[width=0.475\textwidth]{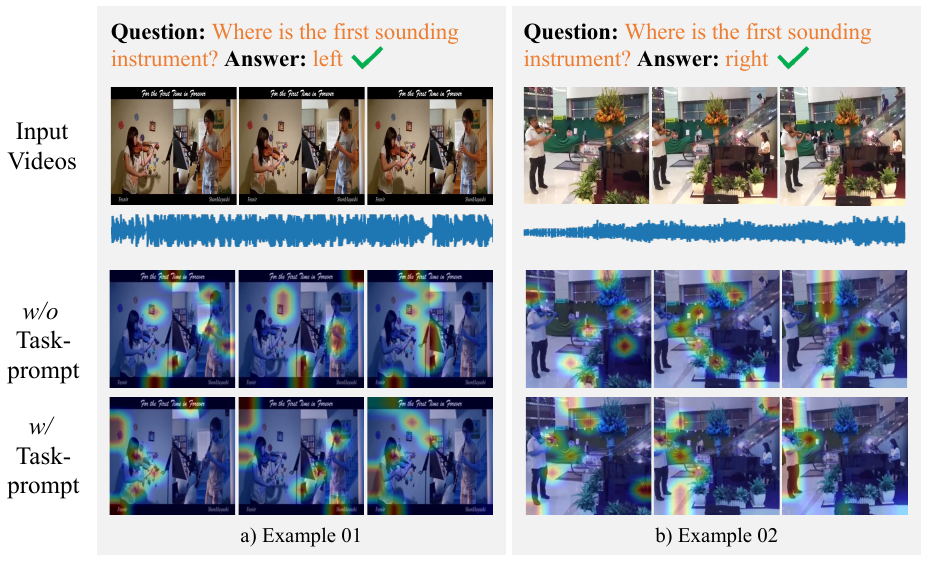}
     \caption{Visualization of the Temporal-Spatial Task (AVQA). The audio-visual representation of the input video is first processed by the AV-Unified framework to obtain a unified representation, corresponding to the \textit{w/o task-specific prompt} setting. Then, a task-specific prompt is applied to guide the model in selecting the most relevant features required for the task, corresponding to the \textit{w/ task-specific prompt} setting.
     }
     \label{fig:vis}
     \vspace{-1em}
\end{figure}

To further analyze the spatiotemporal representations learned by the AV-Unified framework, we visualize the results of the AVQA task using heatmaps. Fig.~\ref{fig:vis} presents two representative examples. These visualizations clearly demonstrate that the task prompt effectively guides the model to focus on task-relevant information within the unified audiovisual representation. Without the task prompt, the model’s attention is scattered and often directed toward irrelevant regions, making it difficult to accurately localize sounding instruments. In contrast, when the prompt is provided, the model attends precisely to the instrument areas, which contain the critical audiovisual cues needed to answer the question correctly. These results highlight the importance of explicit task guidance in improving the model’s ability to extract meaningful spatiotemporal information from complex audio visual scenes.

\vspace{-0.5em}
\subsection{Discussion with Related Works}
Crab and the AV-Unified are proposed around the same period, with Crab capable of jointly handling multiple audiovisual tasks and achieving results of notable reference value. However, the core design philosophies of the two frameworks differ significantly. Crab relies on constructing a large-scale audiovisual dataset to first train a general audiovisual scene understanding model, followed by task-specific fine-tuning. This approach requires substantial additional data and incurs high computational costs. In contrast, AV-Unified is designed based on the intrinsic characteristics of each task. By leveraging the proposed MS-TSPM module, it effectively captures the spatiotemporal correlations in audiovisual signals, achieving strong performance across multiple tasks without extra data or fine-tuning. Furthermore, while Crab’s performance gains primarily depend on the generalization capabilities of the pretrained large model, AV-Unified explicitly models audiovisual spatiotemporal relationships, directly providing essential support for its performance in multi-task scenarios.

\vspace{-0.25em}
\section{Conclustion}

In this paper, we propose AV-Unified, a unified framework that integrates event localization, video parsing, spatial localization, segmentation, and question answering within the context of audiovisual scene understanding. AV-Unified reformulates all tasks into a unified sequence-to-sequence format and trains them using a shared-parameter network for joint learning. To address challenges such as multi-scale temporal events and the lack of supervision linking spatial visual objects with corresponding sounds, we introduce a Multi-scale Spatiotemporal Perception Model (MS-TSPM), which effectively captures events across different temporal scales and models spatial audiovisual associations. Extensive experiments on multiple benchmark datasets validate the effectiveness and robustness of the proposed framework, demonstrating its strong potential for comprehensive audiovisual scene understanding.

Nevertheless, we observe that AV-Unified's performance on certain subtasks remains suboptimal. This is primarily due to the complexity of achieving effective cross-task collaboration, which often requires extensive experimentation and larger-scale data support. 
Future research may address this by incorporating larger and more diverse audiovisual datasets, exploring more advanced architectures, and designing better training strategies.
While the current framework represents an initial step toward unified audiovisual modeling, future work could extend it to broader task sets. Overall, we believe this study lays a solid foundation and offers a promising direction for unified audiovisual scene understanding.

\bibliographystyle{IEEEtran}
\bibliography{main}

% \newpage
\vspace{1em}
\section{Biography Section}
\vspace{-2em}

\begin{IEEEbiography}[{\includegraphics[width=1in,height=1.25in,clip,keepaspectratio]{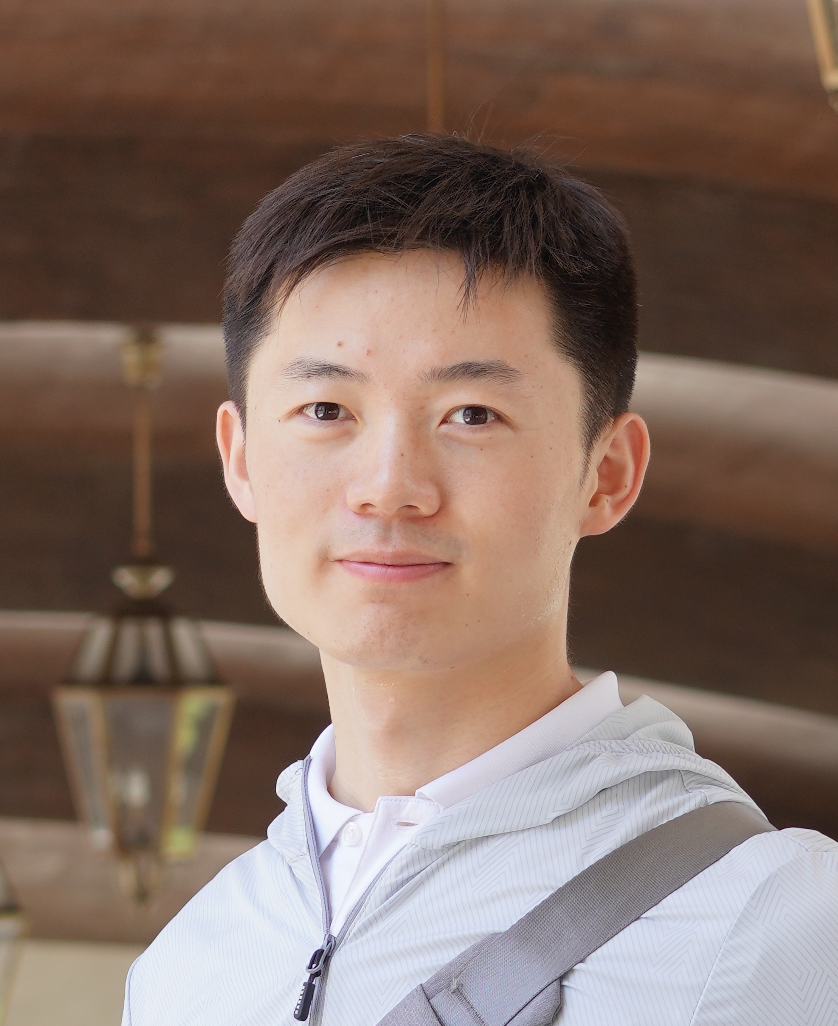}}]{Guangyao Li}
is currently a postdoctoral researcher in the Department of Computer Science and Technology at Tsinghua University. He received his Ph.D. degree from the Gaoling School of Artificial Intelligence, Renmin University of China, in 2024. His research interests include multimodal learning and audio-visual scene understanding.
\vspace{-2em}
\end{IEEEbiography}

\begin{IEEEbiography}[{\includegraphics[width=1in,height=1.25in,clip,keepaspectratio]{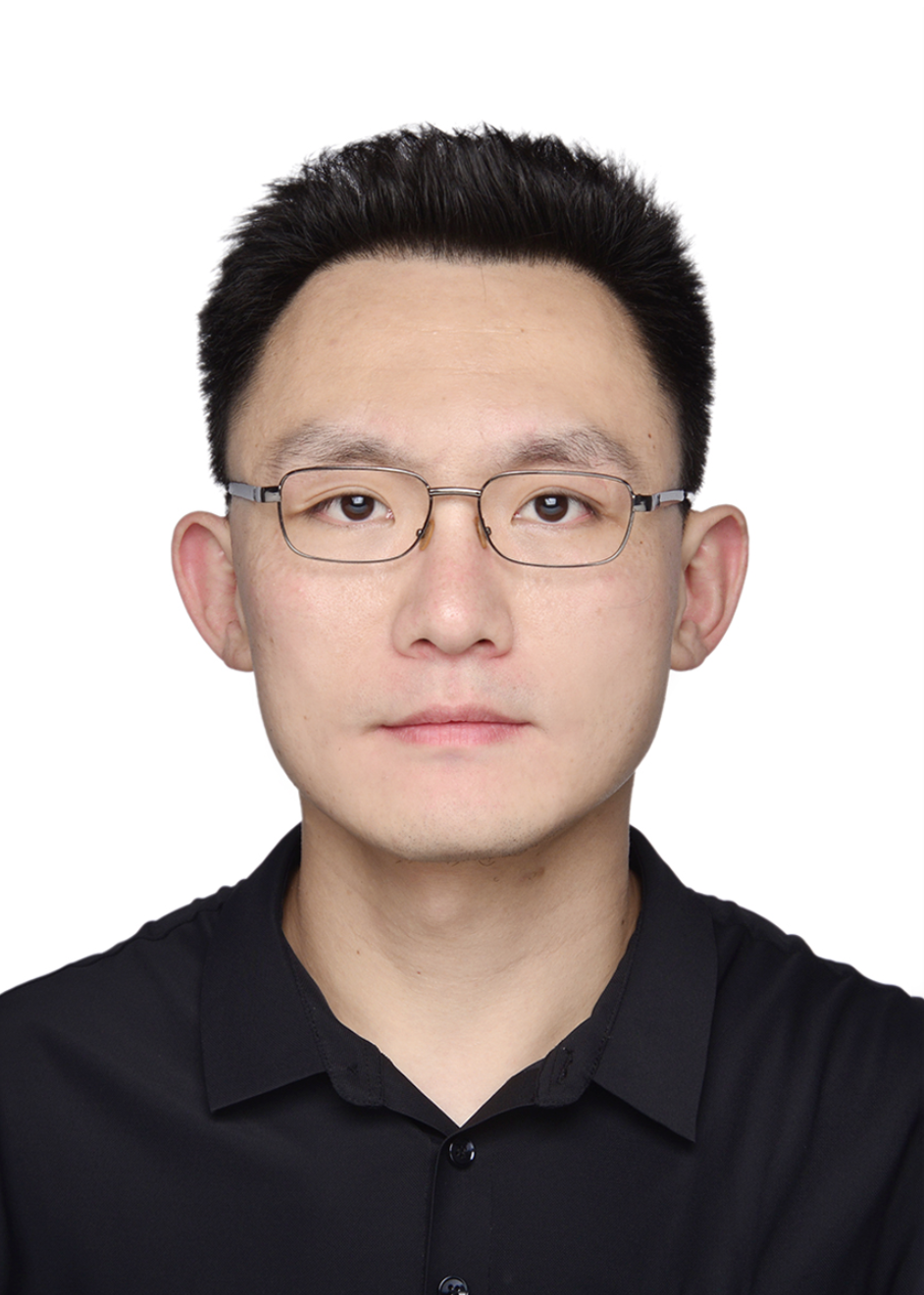}}]{Xin Wang}
is currently an Associate Professor at the Department of Computer Science and Technology, Tsinghua University. He got both of his Ph.D. and B.E degrees in Computer Science and Technology from Zhejiang University, China. He also holds a Ph.D. degree in Computing Science from Simon Fraser University, Canada. His research interests include multimedia intelligence, machine learning and its applications. He has published over 200 highquality research papers in ICML, NeurIPS, IEEE TPAMI, IEEE TKDE, ACM KDD, WWW, ACM SIGIR, ACM Multimedia etc., winning three best paper awards including ACM Multimedia Asia. He is the recipient of ACM China Rising Star Award, IEEE TCMC Rising Star Award and DAMO Academy Young Fellow.
\vspace{-2em}
\end{IEEEbiography}

\begin{IEEEbiography}[{\includegraphics[width=1in,height=1.25in,clip,keepaspectratio]{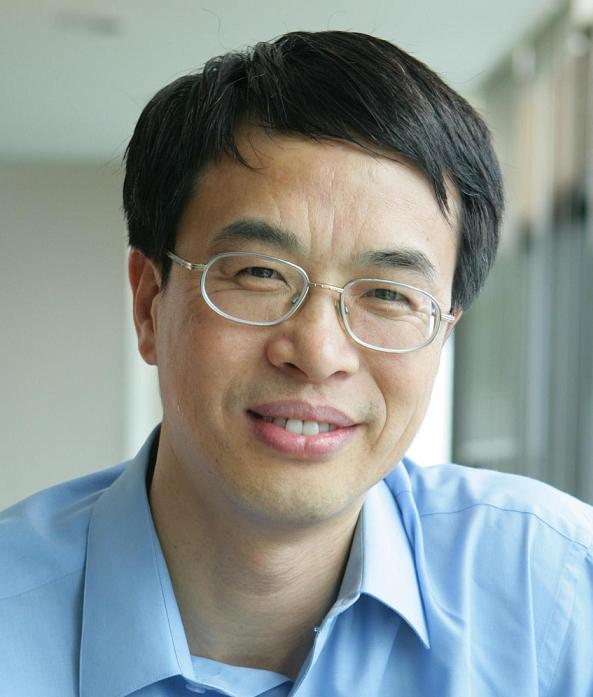}}]{Wenwu Zhu}
is currently a Professor in the Department of Computer Science and Technology at Tsinghua University. He also serves as the Vice Dean of National Research Center for Information Science and Technology, and the Vice Director of Tsinghua Center for Big Data. Prior to his current post, he was a Senior Researcher and Research Manager at Microsoft Research Asia. He was the Chief Scientist and Director at Intel Research China from 2004 to 2008. He worked at Bell Labs, New Jersey as Member of Technical Staff during 1996-1999. He received his Ph.D. degree from New York University in 1996. His research interests are in the area of data-driven multimedia networking and Crossmedia big data computing. He has published over 400 referred papers and is the inventor or co-inventor of over 100 patents. He received eight Best Paper Awards, including ACM Multimedia 2012 and IEEE Transactions on Circuits and Systems for Video Technology in 2001 and 2019.

He served as EiC for IEEE Transactions on Multimedia (2017-2019) and IEEE Transactions on Circuits and Systems for Video Technology (2024-2025). He served in the steering committee for IEEE Transactions on Multimedia (2015-2016) and IEEE Transactions on Mobile Computing (2007-2010), respectively. He serves as General Co-Chair for ACM Multimedia 2018 and ACM CIKM 2019, respectively. He is an AAAS Fellow, IEEE Fellow, SPIE Fellow, and a member of The Academy of Europe (Academia Europaea).
\end{IEEEbiography}

\vfill

\end{document}